# Learning language through pictures

**Grzegorz Chrupała**
g.chrupala@uvt.nl

**Ákos Kádár**
a.kadar@uvt.nl

**Afra Alishahi**
a.alishahi@uvt.nl

Tilburg Center for Cognition and Communication
Tilburg University

## Abstract

We propose IMAGINET, a model of learning visually grounded representations of language from coupled textual and visual input. The model consists of two Gated Recurrent Unit networks with shared word embeddings, and uses a multi-task objective by receiving a textual description of a scene and trying to concurrently predict its visual representation and the next word in the sentence. Mimicking an important aspect of human language learning, it acquires meaning representations for individual words from descriptions of visual scenes. Moreover, it learns to effectively use sequential structure in semantic interpretation of multi-word phrases.

## 1 Introduction

Vision is the most important sense for humans and visual sensory input plays an important role in language acquisition by grounding meanings of words and phrases in perception. Similarly, in practical applications processing multimodal data where text is accompanied by images or videos is increasingly important. In this paper we propose a novel model of learning visually-grounded representations of language from paired textual and visual input. The model learns language through comprehension and production, by receiving a textual description of a scene and trying to "imagine" a visual representation of it, while predicting the next word at the same time.

The full model, which we dub IMAGINET, consists of two Gated Recurrent Unit (GRU) networks coupled via shared word embeddings. IMAGINET uses a multi-task Caruana (1997) objective: both networks read the sentence word-by-word in parallel; one of them predicts the feature representation of the image depicting the described scene after reading the whole sentence, while the other one predicts the next word at each position in the word sequence. The importance of the visual and textual objectives can be traded off, and either of them can be switched off entirely, enabling us to investigate the impact of visual vs textual information on the learned language representations.

Our approach to modeling human language learning has connections to recent models of image captioning (see Section 2). Unlike in many of these models, in IMAGINET the image is the target to predict rather then the input, and the model can build a visually-grounded representation of a sentence independently of an image. We can directly compare the performance of IMAGINET against a simple multivariate linear regression model with bag-of-words features and thus quantify the contribution of the added expressive power of a recurrent neural network.

We evaluate our model's knowledge of word meaning and sentence structure through simulating human judgments of word similarity, retrieving images corresponding to single words as well as full sentences, and retrieving paraphrases of image captions. In all these tasks the model outperforms the baseline; the model significantly correlates with human ratings of word similarity, and predicts appropriate visual interpretations of single and multi-word phrases. The acquired knowledge of sentence structure boosts the model's performance in both image and caption retrieval.

## 2 Related work

Several computational models have been proposed to study early language acquisition. The acquisition of word meaning has been mainly modeled using connectionist networks that learn to associate word forms with semantic or perceptual features (e.g., Li et al., 2004; Coventry et al., 2005; Regier, 2005), and rule-based or probabilistic implementations which use statistical reg-ularities observed in the input to detect associations between linguistic labels and visual features or concepts (e.g., Siskind, 1996; Yu, 2008; Fazly et al., 2010). These models either use toy languages as input (e.g., Siskind, 1996), or child-directed utterances from the CHILDES database (MacWhinney, 2014) paired with artificially generated semantic information. Some models have investigated the acquisition of terminology for visual concepts from simple videos (Fleischman and Roy, 2005; Skocaj et al., 2011). Lazaridou et al. (2015) adapt the skip-gram word-embedding model (Mikolov et al., 2013) for learning word representations via a multi-task objective similar to ours, learning from a dataset where some words are individually aligned with corresponding images. All these models ignore sentence structure and treat inputs as bags of words.

A few models have looked at the concurrent acquisition of words and some aspect of sentence structure, such as lexical categories (Alishahi and Chrupała, 2012) or syntactic properties (Howell et al., 2005; Kwiatkowski et al., 2012), from utterances paired with an artificially generated representation of their meaning. To our knowledge, no existing model has been proposed for concurrent learning of grounded word meanings and sentence structure from large scale data and realistic visual input.

Recently, the engineering task of generating captions for images has received a lot of attention (Karpathy and Fei-Fei, 2014; Mao et al., 2014; Kiros et al., 2014; Donahue et al., 2014; Vinyals et al., 2014; Venugopalan et al., 2014; Chen and Zitnick, 2014; Fang et al., 2014). From the point of view of modeling, the research most relevant to our interests is that of Chen and Zitnick (2014). They develop a model based on a context-dependent recurrent neural network (Mikolov and Zweig, 2012) which simultaneously processes textual and visual input and updates two parallel hidden states. Unlike theirs, our model receives the visual target only at the end of the sentence and is thus encouraged to store in the final hidden state of the visual pathway all aspects of the sentence needed to predict the image features successfully. Our setup is more suitable for the goal of learning representations of complete sentences.

## 3 Models

IMAGINET consists of two parallel recurrent pathways coupled via shared word embeddings. Both pathways are composed of Gated Recurrent Units (GRU) first introduced by Cho et al. (2014) and Chung et al. (2014). GRUs are related to the Long Short-Term Memory units (Hochreiter and Schmidhuber, 1997), but do not employ a separate memory cell. In a GRU, activation at time $t$ is the linear combination of previous activation, and candidate activation:

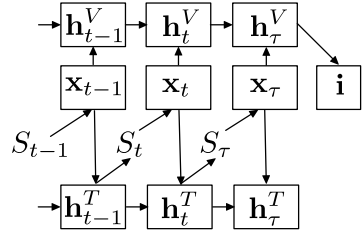

Figure 1: Structure of IMAGINET

$$\mathbf{h}_t = (1 - \mathbf{z}_t) \odot \mathbf{h}_{t-1} + \mathbf{z}_t \odot \tilde{\mathbf{h}}_t \quad (1)$$

where $\odot$ is elementwise multiplication. The update gate determines how much the activation is updated:

$$\mathbf{z}_t = \sigma_s(\mathbf{W}_z \mathbf{x}_t + \mathbf{U}_z \mathbf{h}_{t-1}) \quad (2)$$

The candidate activation is computed as:

$$\tilde{\mathbf{h}}_t = \sigma(\mathbf{W} \mathbf{x}_t + \mathbf{U}(\mathbf{r}_t \odot \mathbf{h}_{t-1})) \quad (3)$$

The reset gate is defined as:

$$\mathbf{r}_t = \sigma_s(\mathbf{W}_r \mathbf{x}_t + \mathbf{U}_r \mathbf{h}_{t-1}) \quad (4)$$

Our gated recurrent units use steep sigmoids for gate activations:

$$\sigma_s(z) = \frac{1}{1 + \exp(-3.75z)}$$

and rectified linear units clipped between 0 and 5 for the unit activations:

$$\sigma(z) = \text{clip}(0.5(z + \text{abs}(z)), 0, 5)$$

Figure 1 illustrates the structure of the network. The word embeddings is a matrix of learned parameters $\mathbf{W}_e$ with each column corresponding to a vector for a particular word. The input word symbol $S_t$ of sentence $S$ at each step $t$ indexes into the embeddings matrix and the vector $\mathbf{x}_t$ forms input to both GRU networks:

$$\mathbf{x}_t = \mathbf{W}_e[:, S_t] \quad (5)$$

This input is mapped into two parallel hidden states, $\mathbf{h}_t^V$ along the visual pathway, and $\mathbf{h}_t^T$ along the textual pathway:

$$\mathbf{h}_t^V = \text{GRU}^V(\mathbf{h}_{t-1}^V, \mathbf{x}_t) \quad (6)$$

$$\mathbf{h}_t^T = \text{GRU}^T(\mathbf{h}_{t-1}^T, \mathbf{x}_t) \quad (7)$$

The final hidden state along the visual pathway $\mathbf{h}_\tau^V$ is then mapped to the predicted target image representation $\hat{\mathbf{i}}$ by the fully connected layer with parameters $\mathbf{V}$ and the clipped rectifier activation:

$$\hat{\mathbf{i}} = \sigma(\mathbf{V}\mathbf{h}_\tau^V) \quad (8)$$

Each hidden state along the textual pathway $\mathbf{h}_t^T$ is used to predict the next symbol in the sentence $S$ via a softmax layer with parameters $\mathbf{L}$:

$$p(S_{t+1}|S_{1:t}) = \text{softmax}(\mathbf{L}\mathbf{h}_t^T) \quad (9)$$

The loss function whose gradient is backpropagated through time to the GRUs and the embeddings is a composite objective with terms penalizing error on the visual and the textual targets simultaneously:

$$L(\theta) = \alpha L^T(\theta) + (1-\alpha)L^V(\theta) \quad (10)$$

where $\theta$ is the set of all IMAGINET parameters. $L^T$ is the cross entropy function:

$$L^T(\theta) = -\frac{1}{\tau}\sum_{t=1}^{\tau} \log p(S_t|S_{1:t}) \quad (11)$$

while $L^V$ is the mean squared error:

$$L^V(\theta) = \frac{1}{K}\sum_{k=1}^{K}(\hat{i}_k - i_k)^2 \quad (12)$$

By setting $\alpha$ to 0 we can switch the whole textual pathway off and obtain the VISUAL model variant. Analogously, setting $\alpha$ to 1 gives the TEXTUAL model. Intermediate values of $\alpha$ (in the experiments below we use 0.1) give the full MULTITASK version. Finally, as baseline for some of the tasks we use a simple linear regression model LINREG with a bag-of-words representation of the sentence:

$$\hat{\mathbf{i}} = \mathbf{A}\mathbf{x} + b \quad (13)$$

where $\hat{\mathbf{i}}$ is the vector of the predicted image features, $\mathbf{x}$ is the vector of word counts for the input sentence and $(\mathbf{A}, b)$ the parameters of the linear model estimated via $L_2$-penalized sum-of-squared-errors loss.

|  | SimLex | MEN 3K |
|---|---|---|
| VISUAL | 0.32 | 0.57 |
| MULTITASK | 0.39 | 0.63 |
| TEXTUAL | 0.31 | 0.53 |
| LINREG | 0.18 | 0.23 |

Table 1: Word similarity correlations with human judgments measured by Spearman's $\rho$ (all correlations are significant at level $p < 0.01$).

## 4 Experiments

**Settings** The model was implemented in Theano (Bastien et al., 2012; Bergstra et al., 2010) and optimized by Adam (Kingma and Ba, 2014).[1] The fixed 4096-dimensional target image representation come from the pre-softmax layer of the 16-layer CNN (Simonyan and Zisserman, 2014). We used 1024 dimensions for the embeddings and for the hidden states of each of the GRU networks. We ran 8 iterations of training, and we report either full learning curves, or the results for each model after iteration 7 (where they performed best for the image retrieval task). For training we use the standard MS-COCO training data. For validation and test, we take a sample of 5000 images each from the validation data.

### 4.1 Word representations

We assess the quality of the learned embeddings for single words via two tasks: (i) we measure similarity between embeddings of word pairs and compare them to elicited human ratings; (ii) we examine how well the model learns visual representations of words by projecting word embeddings into the visual space, and retrieving images of single concepts from ImageNet.

**Word similarity judgment** For similarity judgment correlations, we selected two existing benchmarks that have the largest vocabulary overlap with our data: MEN 3K (Bruni et al., 2014) and SimLex-999 (Hill et al., 2014). We measure the similarity between word pairs by computing the cosine similarity between their embeddings from three versions of our model, VISUAL, MULTITASK and TEXTUAL, and the baseline LINREG.

Table 1 summarizes the results. All IMAGINET models significantly correlate with human similarity judgments, and outperform LINREG. Examples of word pairs for which MULTITASK cap-tures human similarity judgments better than VISUAL include antonyms (*dusk*, *dawn*), collocations (*sexy*, *smile*), or related but not visually similar words (*college*, *exhibition*).

---
[1] Code available at github.com/gchrupala/imaginet.

| VISUAL | MULTITASK | LINREG |
|--------|-----------|--------|
| 0.38 | 0.38 | 0.33 |

Table 2: Accuracy@5 of retrieving images with compatible labels from ImageNet.

**Single-word image retrieval** In order to visualize the acquired meaning for individual words, we use images from the ILSVRC2012 subset of ImageNet (Russakovsky et al., 2014) as benchmark. Labels of the images in ImageNet are synsets from WordNet, which identify a single concept in the image rather than providing descriptions of its full content. Since the synset labels in ImageNet are much more precise than the descriptions provided in the captions in our training data (e.g., *elkhound*), we use synset hypernyms from WordNet as substitute labels when the original labels are not in our vocabulary.

We extracted the features from the 50,000 images of the ImageNet validation set. The labels in this set result in 393 distinct (original or hypernym) words from our vocabulary. Each word was projected to the visual space by feeding it through the model as a one-word sentence. We ranked the vectors corresponding to all 50,000 images based on their similarity to the predicted vector, and measured the accuracy of retrieving an image with the correct label among the top 5 ranked images (Accuracy@5). Table 2 summarizes the results: VISUAL and MULTITASK learn more accurate word meaning representations than LINREG.

### 4.2 Sentence structure

In the following experiments, we examine the knowledge of sentence structure learned by IMAGINET, and its impact on the model performance on image and paraphrase retrieval.

**Image retrieval** We retrieve images based on the similarity of their vectors with those predicted by IMAGINET in two conditions: sentences are fed to the model in their original order, or scrambled. Figure 2 (left) shows the proportion of sentences for which the correct image was in the top 5 highest ranked images for each model, as a function of the number of training iterations: both models out-perform the baseline. MULTITASK is initially better in retrieving the correct image, but eventually the gap disappears. Both models perform substantially better when tested on the original captions compared to the scrambled ones, indicating that models learn to exploit aspects of sentence structure. This ability is to be expected for MULTITASK, but the VISUAL model shows a similar effect to some extent. In the case of VISUAL, this sensitivity to structural aspects of sentence meaning is entirely driven by how they are reflected in the image, as this models only receives the visual supervision signal.

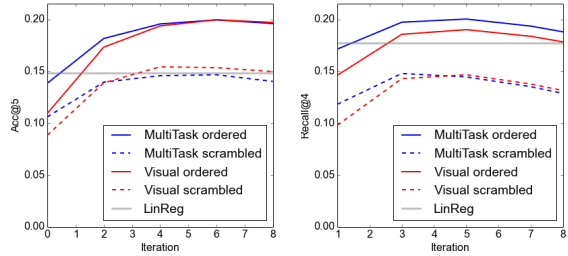

Figure 2: Left: Accuracy@5 of **image retrieval** with original versus scrambled captions. Right: Recall@4 of **paraphrase retrieval** with original vs scrambled captions.

Qualitative analysis of the role of sequential structure suggests that the models are sensitive to the fact that periods terminate a sentence, that sentences tend not to start with conjunctions, that topics appear in sentence-initial position, and that words have different importance as modifiers versus heads. Figure 3 shows an example; see supplementary material for more.

**IMAGINET vs captioning systems** While it is not our goal to engineer a state-of-the-art image retrieval system, we want to situate IMAGINET's performance within the landscape of image retrieval results on captioned images. As most of these are on Flickr30K (Young et al., 2014), we ran MULTITASK on it and got an accuracy@5 of 32%, within the range of numbers reported in previous work: 29.8% (Socher et al., 2014), 31.2% (Mao et al., 2014), 34% (Kiros et al., 2014) and 37.7% (Karpathy and Fei-Fei, 2014). Karpathy and Fei-Fei (2014) report 29.6% on MS-COCO, but with additional training data.

| | |
|---|---|
| Original | a couple of horses UNK their head over a rock pile |
| rank 1 | two brown horses hold their heads above a rocky wall . |
| rank 2 | two horses looking over a short stone wall . |
| Scrambled | rock couple their head pile a a UNK over of horses |
| rank 1 | an image of a man on a couple of horses |
| rank 2 | looking in to a straw lined pen of cows |
| Original | a cute baby playing with a cell phone |
| rank 1 | small baby smiling at camera and talking on phone . |
| rank 2 | a smiling baby holding a cell phone up to ear . |
| Scrambled | phone playing cute cell a with baby a |
| rank 1 | someone is using their phone to send a text or play a game . |
| rank 2 | a camera is placed next to a cellular phone . |

Table 3: Examples of two nearest neighbors retrieved by MULTITASK for original and scrambled captions.

*" a variety of kitchen utensils hanging from a UNK board ."*

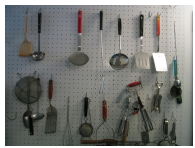 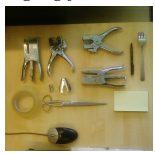

*"kitchen of from hanging UNK variety a board utensils a ."*

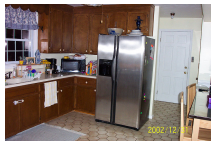 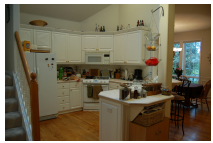

Figure 3: For the original caption MULTITASK understands *kitchen* as a modifier of headword *utensils*, which is the topic. For the scrambled sentence, the model thinks *kitchen* is the topic.

**Paraphrase retrieval** In our dataset each image is paired with five different captions, which can be seen as paraphrases. This affords us the opportunity to test IMAGINET's sentence representations on a non-visual task. Although all models receive one caption-image pair at a time, the co-occurrence with the same image can lead the model to learn structural similarities between captions that are different on the surface. We feed the whole set of validation captions through the trained model and record the final hidden visual state $\mathbf{h}_\tau^V$. For each caption we rank all others according to cosine similarity and measure the proportion of the ones associated with the same image among the top four highest ranked. For the scrambled condition, we rank original captions against a scrambled one. Figure 2 (right) summarizes the results: both models outperform the baseline on ordered captions, but not on scrambled ones. As expected, MULTITASK is more affected by manipulating word order, because it is more sensitive to structure. Table 3 shows concrete examples of the effect of scrambling words in what sentences are retrieved.

## 5 Discussion

IMAGINET is a novel model of grounded language acquisition which simultaneously learns word meaning representations and knowledge of sentence structure from captioned images. It acquires meaning representations for individual words from descriptions of visual scenes, mimicking an important aspect of human language learning, and can effectively use sentence structure in semantic interpretation of multi-word phrases. In future we plan to upgrade the current word-prediction pathway to a sentence reconstruction and/or sentence paraphrasing task in order to encourage the formation of representations of full sentences. We also want to explore the acquired structure further, especially for generalizing the grounded meanings to those words for which visual data is not available.

## Acknowledgements

The authors would like to thank Angeliki Lazaridou and Marco Baroni for their many insightful comments on the research presented in this paper.

## References

Afra Alishahi and Grzegorz Chrupała. 2012. Concurrent acquisition of word meaning and lexical categories. In *Proceedings of the 2012 Joint Conference on Empirical Methods in Natural Language Processing and Computational Natural Language Learning*, pages 643–654. Association for Computational Linguistics.

Frédéric Bastien, Pascal Lamblin, Razvan Pascanu, James Bergstra, Ian J. Goodfellow, Arnaud Berg-

eron, Nicolas Bouchard, and Yoshua Bengio. 2012. Theano: new features and speed improvements. Deep Learning and Unsupervised Feature Learning NIPS 2012 Workshop.

James Bergstra, Olivier Breuleux, Frédéric Bastien, Pascal Lamblin, Razvan Pascanu, Guillaume Desjardins, Joseph Turian, David Warde-Farley, and Yoshua Bengio. 2010. Theano: a CPU and GPU math expression compiler. In *Proceedings of the Python for Scientific Computing Conference (SciPy)*. Oral Presentation.

Elia Bruni, Nam-Khanh Tran, and Marco Baroni. 2014. Multimodal distributional semantics. *Journal of Artificial Intelligence Research (JAIR)*, 49:1–47.

Rich Caruana. 1997. Multitask learning. *Machine learning*, 28(1):41–75.

Xinlei Chen and C Lawrence Zitnick. 2014. Learning a recurrent visual representation for image caption generation. *arXiv preprint arXiv:1411.5654*.

Kyunghyun Cho, Bart van Merriënboer, Dzmitry Bahdanau, and Yoshua Bengio. 2014. On the properties of neural machine translation: Encoder-decoder approaches. In *Eighth Workshop on Syntax, Semantics and Structure in Statistical Translation (SSST-8)*.

Junyoung Chung, Caglar Gulcehre, KyungHyun Cho, and Yoshua Bengio. 2014. Empirical evaluation of gated recurrent neural networks on sequence modeling. In *NIPS 2014 Deep Learning and Representation Learning Workshop*.

Kenny R. Coventry, Angelo Cangelosi, Rohanna Rajapakse, Alison Bacon, Stephen Newstead, Dan Joyce, and Lynn V. Richards. 2005. Spatial prepositions and vague quantifiers: Implementing the functional geometric framework. In Christian Freksa, Markus Knauff, Bernd Krieg-Brückner, Bernhard Nebel, and Thomas Barkowsky, editors, *Spatial Cognition IV. Reasoning, Action, Interaction*, volume 3343 of *Lecture Notes in Computer Science*, pages 98–110. Springer Berlin Heidelberg.

Jeff Donahue, Lisa Anne Hendricks, Sergio Guadarrama, Marcus Rohrbach, Subhashini Venugopalan, Kate Saenko, and Trevor Darrell. 2014. Long-term recurrent convolutional networks for visual recognition and description. *arXiv preprint arXiv:1411.4389*.

Hao Fang, Saurabh Gupta, Forrest Iandola, Rupesh Srivastava, Li Deng, Piotr Dollár, Jianfeng Gao, Xiaodong He, Margaret Mitchell, John Platt, et al. 2014. From captions to visual concepts and back. *arXiv preprint arXiv:1411.4952*.

Afsaneh Fazly, Afra Alishahi, and Suzanen Stevenson. 2010. A probabilistic computational model of cross-situational word learning. *Cognitive Science: A Multidisciplinary Journal*, 34(6):1017–1063.

Michael Fleischman and Deb Roy. 2005. Intentional context in situated natural language learning. In *Proceedings of the Ninth Conference on Computational Natural Language Learning*, pages 104–111. Association for Computational Linguistics.

Felix Hill, Roi Reichart, and Anna Korhonen. 2014. Simlex-999: Evaluating semantic models with (genuine) similarity estimation. *arXiv preprint arXiv:1408.3456*.

Sepp Hochreiter and Jürgen Schmidhuber. 1997. Long short-term memory. *Neural computation*, 9(8):1735–1780.

Steve R Howell, Damian Jankowicz, and Suzanna Becker. 2005. A model of grounded language acquisition: Sensorimotor features improve lexical and grammatical learning. *Journal of Memory and Language*, 53(2):258–276.

Andrej Karpathy and Li Fei-Fei. 2014. Deep visual-semantic alignments for generating image descriptions. *arXiv preprint arXiv:1412.2306*.

Diederik P. Kingma and Jimmy Ba. 2014. Adam: A method for stochastic optimization. *CoRR*, abs/1412.6980.

Ryan Kiros, Ruslan Salakhutdinov, and Richard S Zemel. 2014. Unifying visual-semantic embeddings with multimodal neural language models. *arXiv preprint arXiv:1411.2539*.

Tom Kwiatkowski, Sharon Goldwater, Luke Zettlemoyer, and Mark Steedman. 2012. A probabilistic model of syntactic and semantic acquisition from child-directed utterances and their meanings. In *Proceedings of the 13th Conference of the European Chapter of the Association for Computational Linguistics*, pages 234–244. Association for Computational Linguistics.

Angeliki Lazaridou, Nghia The Pham, and Marco Baroni. 2015. Combining language and vision with a multimodal skip-gram model. In *Proceedings of NAACL HLT 2015 (2015 Conference of the North American Chapter of the Association for Computational Linguistics - Human Language Technologies)*.

Ping Li, Igor Farkas, and Brian MacWhinney. 2004. Early lexical development in a self-organizing neural network. *Neural Networks*, 17:1345–1362.

Brian MacWhinney. 2014. *The CHILDES project: Tools for analyzing talk, Volume I: Transcription format and programs*. Psychology Press.

Junhua Mao, Wei Xu, Yi Yang, Jiang Wang, and Alan L Yuille. 2014. Explain images with multimodal recurrent neural networks. In *NIPS 2014 Deep Learning Workshop*.

Tomas Mikolov, Ilya Sutskever, Kai Chen, Greg S Corrado, and Jeff Dean. 2013. Distributed representations of words and phrases and their compositional-ity. In *Advances in Neural Information Processing Systems*, pages 3111–3119.

Tomas Mikolov and Geoffrey Zweig. 2012. Context dependent recurrent neural network language model. In *SLT*, pages 234–239.

Terry Regier. 2005. The emergence of words: Attentional learning in form and meaning. *Cognitive Science: A Multidisciplinary Journal*, 29:819–865.

Olga Russakovsky, Jia Deng, Hao Su, Jonathan Krause, Sanjeev Satheesh, Sean Ma, Zhiheng Huang, Andrej Karpathy, Aditya Khosla, Michael Bernstein, Alexander C. Berg, and Li Fei-Fei. 2014. ImageNet Large Scale Visual Recognition Challenge.

K. Simonyan and A. Zisserman. 2014. Very deep convolutional networks for large-scale image recognition. *CoRR*, abs/1409.1556.

Jeffrey M. Siskind. 1996. A computational study of cross-situational techniques for learning word-to-meaning mappings. *Cognition*, 61(1-2):39–91.

Danijel Skocaj, Matej Kristan, Alen Vrecko, Marko Mahnic, Miroslav Janicek, Geert-Jan M Kruijff, Marc Hanheide, Nick Hawes, Thomas Keller, Michael Zillich, et al. 2011. A system for interactive learning in dialogue with a tutor. In *Intelligent Robots and Systems (IROS), 2011 IEEE/RSJ International Conference on*, pages 3387–3394. IEEE.

Richard Socher, Andrej Karpathy, Quoc V Le, Christopher D Manning, and Andrew Y Ng. 2014. Grounded compositional semantics for finding and describing images with sentences. *Transactions of the Association for Computational Linguistics*, 2:207–218.

Subhashini Venugopalan, Huijuan Xu, Jeff Donahue, Marcus Rohrbach, Raymond Mooney, and Kate Saenko. 2014. Translating videos to natural language using deep recurrent neural networks. *arXiv preprint arXiv:1412.4729*.

Oriol Vinyals, Alexander Toshev, Samy Bengio, and Dumitru Erhan. 2014. Show and tell: A neural image caption generator. *arXiv preprint arXiv:1411.4555*.

Peter Young, Alice Lai, Micah Hodosh, and Julia Hockenmaier. 2014. From image descriptions to visual denotations: New similarity metrics for semantic inference over event descriptions. *Transactions of the Association for Computational Linguistics*, 2:67–78.

Chen Yu. 2008. A statistical associative account of vocabulary growth in early word learning. *Language Learning and Development*, 4(1):32–62.

## A  Image retrieval with single words

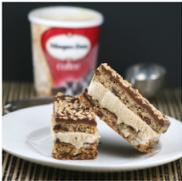 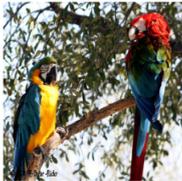

| Keyword: | *dessert* | *parrot* |
| Original label: | *ice cream* | *macaw* |
| Hypernym: | *dessert* | *parrot* |

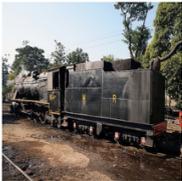 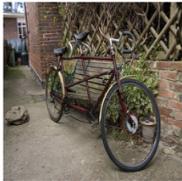

| Keyword: | *locomotive* | *bicycle* |
| Original label: | *steam locomotive* | *bicycle-built-for-two* |
| Hypernym: | *locomotive* | *bicycle* |

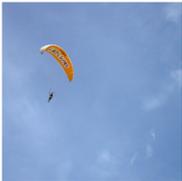 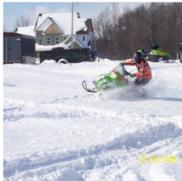

| Keyword: | *parachute* | *snowmobile* |
| Original label: | *parachute* | *snowmobile* |

Figure 4: Sample images for single words. Under the images are the keywords that were used for the retrieval, the original label of the images and if it was not in our vocabulary its hypernym is included.

We visualize the acquired meaning of individual words using images from the ILSVRC2012 subset of ImageNet (Russakovsky et al., 2014). Labels of the images in ImageNet are synsets from WordNet, which identify a single concept in the image rather than providing descriptions of its full content. When the synset labels in ImageNet are too specific and cannot be found in our vocabulary, we replace them with their hypernyms from WordNet.

Figure 4 shows examples of images retrieved via projections of single words into the visual space using the MULTITASK model. As can be seen, the predicted images are intuitive. For those for which we use the hypernym as key, the more general term (e.g. *parrot*) is much more common in humans' daily descriptions of visual scenes than the original label used in ImageNet (e.g. *macaw*). The quantitative evaluation of this task is reported in the body of the paper.

## B  Effect of scrambling word order

In Figures 5–7 we show some illustrative cases of the effect for image retrieval of scrambling the input captions to the MULTITASK model trained on un-scrambled ones. These examples suggest that the model learns a number of facts about sentence structure. They range from very obvious, e.g. periods terminate sentences, to quite interesting, such as the distinction between modifiers and heads or the role of word order in encoding information structure (i.e. the distinction between topic and comment).

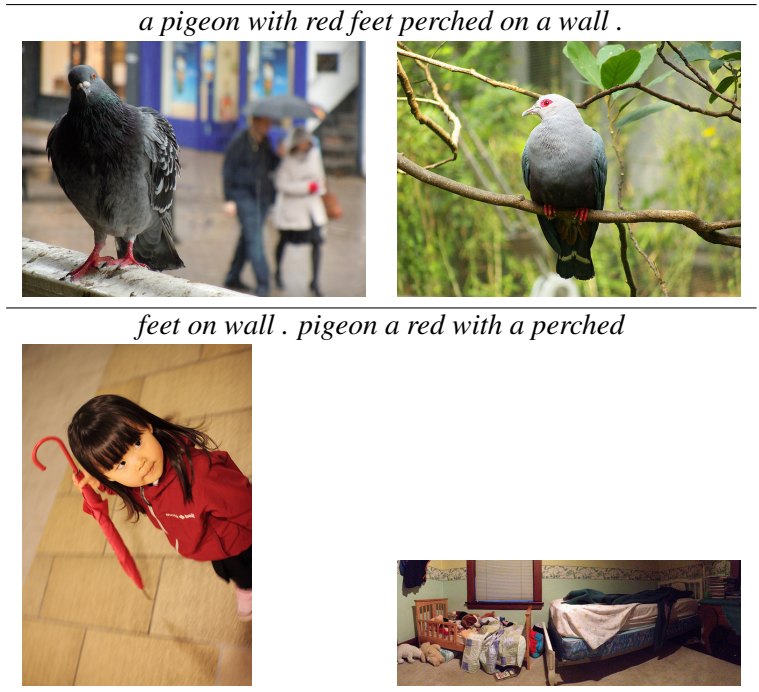

*a pigeon with red feet perched on a wall .*

*feet on wall . pigeon a red with a perched*

Figure 5: In the scrambled sentence, the presence of a full stop in the middle of a sentence causes all material following it to be ignored, so the model finds pictures with wall-like objects.

## C  Propagating distributional information through Multi-Task objective

Table 4 lists example word pairs for which the MULTITASK model matches human judgments closer than the VISUAL model. Some interesting cases are words which are closely related but which have the opposite meaning (*dawn, dusk*), or words which denote entities from the same broad class, but which are visually very dissimilar (*insect, lizard*). There are, however, also examples where there is no obvious prior expectation for the MULTITASK model to do better, e.g. (*maple, oak*).

| Word 1 | Word 2 | Human | MULTITASK | VISUAL |
|---|---|---|---|---|
| construction | downtown | 0.5 | 0.5 | 0.2 |
| sexy | smile | 0.4 | 0.4 | 0.2 |
| dawn | dusk | 0.8 | 0.7 | 0.4 |
| insect | lizard | 0.6 | 0.5 | 0.2 |
| dawn | sunrise | 0.9 | 0.7 | 0.4 |
| collage | exhibition | 0.6 | 0.4 | 0.2 |
| bikini | swimsuit | 0.9 | 0.7 | 0.4 |
| outfit | skirt | 0.7 | 0.5 | 0.2 |
| sun | sunlight | 1.0 | 0.7 | 0.4 |
| maple | oak | 0.9 | 0.5 | 0.2 |
| shirt | skirt | 0.9 | 0.4 | 0.1 |

Table 4: A sample of word pairs from the MEN 3K dataset for which the MULTITASK model matches human judgments better than VISUAL. All scores are scaled to the [0, 1] range.

*blue and silver motorcycle parked on pavement under plastic awning .*

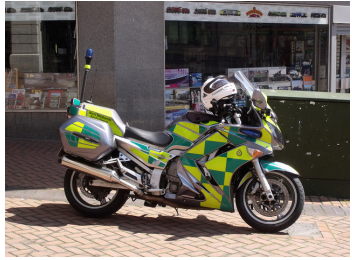
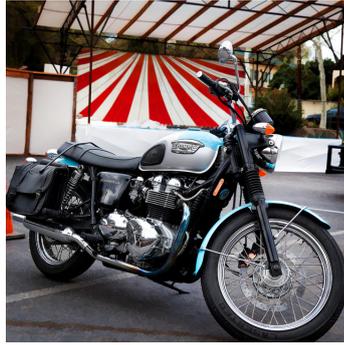

*pavement silver awning and motorcycle blue on under plastic . parked*

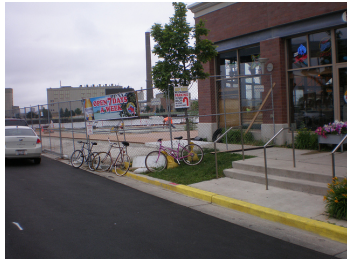
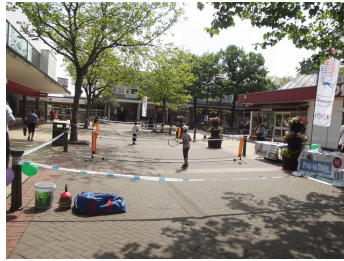

Figure 6: The model understands that *motorcycle* is the topic, even though it's not the very first word. In the scrambled sentence is treats *pavement* as the topic.

*a brown teddy bear laying on top of a dry grass covered ground .*

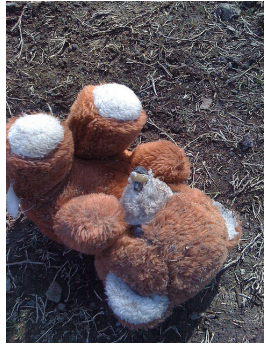
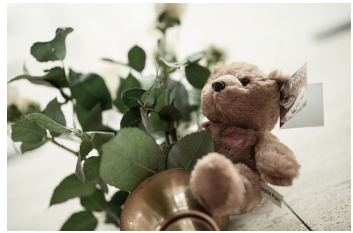

*a a of covered laying bear on brown grass top teddy ground . dry*

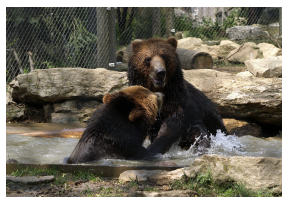
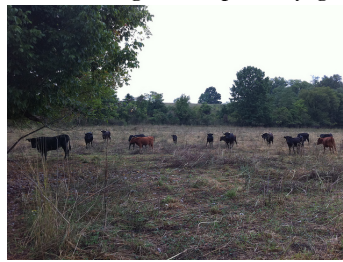

Figure 7: The model understands the compound *teddy bear*. In the scrambled sentence, it finds a picture of real bears instead.